\pdfoutput=1

\documentclass[11pt]{article}

\usepackage{acl}

\usepackage{times}
\usepackage{latexsym}

\usepackage[T1]{fontenc}

\usepackage[utf8]{inputenc}

\usepackage{microtype}

\usepackage{graphicx}
\usepackage{amsmath}
\usepackage{amsfonts,amssymb}
\usepackage{multirow}
\usepackage{booktabs}
\usepackage{subcaption}
\usepackage{xcolor}

\newcommand{\thickhline}{\noalign{\hrule height 1pt}}

\title{\textsc{TANet}: Thread-Aware Pretraining for Abstractive Conversational Summarization}

\author{
  Ze Yang$^1$, Liran Wang$^1$, Zhoujin Tian$^1$, Wei Wu$^2$, \textbf{Zhoujun Li}$^1$\thanks{~~~Corresponding Author}~~~~~\\
  $^1$State Key Lab of Software Development Environment, Beihang University, Beijing, China\\
  $^2$Meituan, Beijing, China \\
  \texttt{\{tobey,wanglr,eitbar,lizj\}@buaa.edu.cn}\\
  \texttt{wuwei19850318@gmail.com}
}

\begin{document}
\maketitle
\begin{abstract}
Although pre-trained language models (PLMs) have achieved great success and become a milestone in NLP, abstractive conversational summarization remains a challenging but less studied task. The difficulty lies in two aspects. One is the lack of large-scale conversational summary data. Another is that applying the existing pre-trained models to this task is tricky because of the structural dependence within the conversation and its informal expression, etc.
In this work, 
we first build a large-scale (11M) pretraining dataset called \textsc{RCS}, based on the multi-person discussions in the Reddit community.
We then present \textsc{TANet}, a thread-aware Transformer-based network. Unlike the existing pre-trained models that treat a conversation as a sequence of sentences, we argue that the inherent contextual dependency among the utterances plays an essential role in understanding the entire conversation and thus propose two new techniques to incorporate the structural information into our model. 
The first is \textit{thread-aware attention} which is computed by taking into account the contextual dependency within utterances. Second, we apply \textit{thread prediction} loss to predict the relations between utterances. 
We evaluate our model on four datasets of real conversations, covering types of meeting transcripts, customer-service records, and forum threads. Experimental results demonstrate that \textsc{TANET} achieves a new state-of-the-art in terms of both automatic evaluation and human judgment.
\end{abstract}

\section{Introduction}
\label{sec:intro}
Text summarization is a long-standing challenging task in artificial intelligence, aiming to condense a piece of text to a shorter version, retaining the critical information.
There are various promising applications of conversational summarization in the real world, emphasizing the need to build auto summarization systems.
For example, online customer-service staff can improve work efficiency by recording the customer demands and current solutions after each communication.
In the industry, meeting summaries are also generally required in order to track the progress of projects. 
The automatic doctor-patient interaction summary can save doctors much time for filling out medical records. 
Therefore, conversational summarization has been a potential field in summarization and has received increasing attention.

\begin{figure}[t!]
    \centering
    \includegraphics[width=\columnwidth]{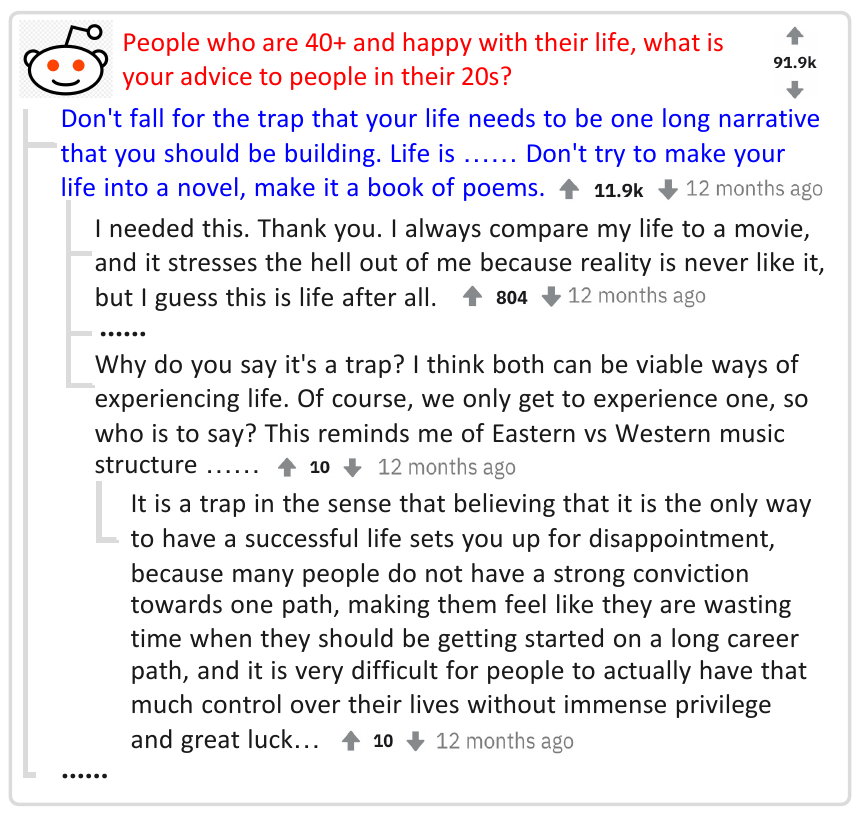}
    \caption{An abbreviated example from corpus RCS.
        It contains a total of 14k comments and more than 210k words in this post. The \textcolor{red}{title} and the \textcolor{blue}{lead comment} are selected as the pseudo summary of this thread.
    }
    \label{fig:corpus}
\end{figure}

Benefiting from the availability of large-scale high-quality data, the abstractive document summarization has been extensively explored in the past years \cite{rush-etal-2015-neural,see-etal-2017-get,chen-bansal-2018-fast}. Recently, the pretraining methods further extend the success \cite{lewis-etal-2020-bart,zhang2020pegasus}. 
In contrast, abstractive conversational summarization is a more challenging but less studied task. 
The reason mainly lies in: 
(1) compared with news, there are no large-scale publicly available labeled datasets for abstractive conversational summarization;
(2) conversations are usually informal, verbose, and repetitive, sprinkled with false starts, backchanneling, reconfirmations, hesitations, and speaker interruptions \cite{sacks1978simplest}, which makes the whole session difficult to understand; 
(3) unlike the linear relationship in the one-speaker document, there are always multiple speakers in a conversation, and the inherent contextual relationship are structured;
(4) conversations in some scenarios could be much longer than a document. For instance, in CNN/Daily Mail dataset \cite{hermann2015teaching}, the average number of words in a document is $781$, while the average length of the transcripts in ICSI, a widely explored meeting corpus, is $10,189$. These challenges encourage us to explore conversation-oriented summarization methods.

To overcome the challenges, we study pretraining for abstractive conversational summarization in this work.
To tackle the bottleneck of insufficient data, we first build a large-scale (11M) corpus for conversational summarization called RCS, based on the multi-person discussions crawled from Reddit website.
Figure \ref{fig:corpus} shows an abbreviated example in RCS. 
For the model architecture, we present \textsc{TANet}, a Thread-Aware NETwork for abstractive conversational summarization. 
As conversations are usually lengthy, we adopt the hierarchical encoders, which consist of a token encoder and an utterance encoder. 
Unlike the existing pre-trained models that treat a conversation as a sequence of sentences, we argue that the inherent contextual dependency among the utterances plays an essential role in understanding the entire conversation and thus propose two new techniques to incorporate the structural information into \textsc{TANet}.
First, we replace the self-attention layers in the utterance encoder with the thread-aware relative attention. Second, we propose a new pretraining task, the thread prediction, to further enhance the representations by predicting the relations across a small set of utterance.

We evaluate \textsc{TANet} on four datasets of conversational summarization, covering domains of meeting transcripts \cite{10.1007/11677482_3,1198793}, customer-service records \cite{yuan2019abstractive} and forum threads \cite{DBLP:conf/flairs/TarnpradabLH17}. Experimental results indicate that \textsc{TANet} achieves new state-of-the-art on all datasets in terms of both automatic evaluation and human judgement.

In summary, our contributions in this work are three-fold: (1) We build a large-scale pretraining corpus based on real conversations for abstractive conversational summarization.
(2) \textsc{TANet} is the first pre-trained abstractive conversational summarization model with inherent structure modeling.
(3) The effectiveness of \textsc{TANet} are demonstrated on four downstream datasets of conversational summarization, covering types of meeting transcripts, customer-service records, and forum threads.

\section{Reddit for Conversational Summarization}
Existing conversational summarization corpora \cite{10.1007/11677482_3,1198793,DBLP:conf/flairs/TarnpradabLH17,yuan2019abstractive,gliwa-etal-2019-samsum} have a low number of conversations, which prevents research community from engaging into this problem.
In this work, to benefit from the large-scale conversation corpus, we mined and processed a large-scale dataset from \textbf{R}eddit \footnote{\url{https://www.reddit.com}} for \textbf{C}onversational \textbf{S}ummarization called \textbf{RCS}. Figure \ref{fig:corpus} shows an example in the dataset. To our best knowledge, RCS is the first large-scale pretraining corpus with real conversations for abstractive conversational summarization.

We crawled the posts on the Reddit site from $2019$ till $2020$. A post is composed of a title and its corresponding discussions, which usually consist of multiple threads. The comments in a thread can naturally expand into a tree structure.
Remarkably, each comment has rich attributes, including the user information, creation timestamp and the accumulated score\footnote{The score is the number of upvotes minus the number of downvotes.}, etc.
With the large-scale real multi-person conversation data, the key is how to construct a summary-like instance for a thread.
We consider two strategies to select sentences that appear to dominate the thread: (1) \textbf{Title}. The discussions of each post are all developed upon the topic of the title, so we select the title as a part of the pseudo summary;
(2) \textbf{Lead Comment}.
Despite the topic given by the title, the lead comment (i.e., the first comment of a thread) also well influences the future direction of what is discussed in this thread.
We concatenate them as the pseudo summary of the discussions in a thread. Lead comment's original position is replaced by a special token $\mathtt{[MASK]}$.

To clean up RCS, we adopted a series of heuristics including: (1) We removed any threads where the number of comments less than $10$; 
(2) We discarded any not-safe-for-work posts, such as posts containing adult or violent content;
(3) We replaced all URLs with token $\mathtt{[URL]}$; 
(4) We removed all markup and any other non-text content such as ``*, $\sim$, [, ]''; 
(5) We removed any threads whose title or lead comment scored less than $0$.
We removed any posts which contain quarantine, picture, or video, etc. 
After then, the dataset has $11,200,981$ instances.

\section{Methodology}
\label{sec:method}
In this section, we present \textsc{TANET}, a thread-aware pretrained model which incorporates the inherent dependencies between utterances to enable improved conversation's representations for summary generation.
Below, we first introduce the model architecture, thread-aware attention, and then introduce our pretraining objectives. Finally, we move on to the application of downstream tasks.

\subsection{Model Architecture}
\paragraph{Encoder.} We employ hierarchical encoders, a \textit{token encoder} and an \textit{utterance encoder}, to represent the input conversation. This design mainly comes from two considerations: (1) The conversations in actual applications are lengthy (e.g., The Reddit post in Figure \ref{fig:corpus} has more than $210$k words, and meeting transcripts usually consists of thousands of tokens.), thus it may not be feasible to simply apply the canonical transformer structure. (2) Hierarchical architecture is more suitable for the conversational tasks to carry out modeling of utterances and interactive structure of the conversation.

Let $C=(u_0,\cdots, u_{|C|})$ denote an conversation instance in the pretraining corpus $\mathcal{D}_p$. $u_i=(\mathtt{\langle bos \rangle}, w_{i,1},\cdots, w_{i,|u_i|})$ is the token sequence of $i$-$th$ utterance after tokenization, where $\mathtt{\langle bos \rangle}$ is a special token in vocabulary $\mathcal{V}$ to represent the beginning of a turn. 
The token encoder takes each sequence $u_i$ as the input and first converts it into input vectors $\mathbf{H}^{\mathcal{T}, 0}_i \in \mathbb{R}^{|u_i|\times d_{h}}$. For each token, its input vector is constructed by summing up the corresponding token embedding and the sine-cosine positional embedding \cite{vaswani2017attention}. Then, $N$ identical layers are nested over $\mathbf{H}^{\mathcal{T}, 0}_i$ to produce the contextual representations by:
\begin{equation}\small
    \label{eq:token-encoder}
    \mathbf{H}^{\mathcal{T},N}_i = \mathrm{Transformer}^{\mathcal{T}}(\mathbf{H}^{\mathcal{T},0}_i)
\end{equation}
Each layer consists of two sub-layers, a self-attention sub-layer followed by a position-wise feed-forward sub-layer and uses residual connections around each of them. We adopt the pre-layer normalization following several recent works \cite{baevski2018adaptive,child2019generating,wang2019learning,xiong2020layer}, which places the layer normalization inside the residual connection. That is, given input $x$, the output of each sub-layer is $x+\mathrm{Sublayer}(\mathrm{LayerNorm}(x))$. 
The utterance encoder also has $N$ identical transformer layers in structure, which processes the information at turn level. All utterances are arranged in the order of their timestamps, and we employ the sine-cosine positional embedding to model the chronological order.
Let $\mathbf{H}^{\mathcal{U},0}=(h_{u_0},\cdots, h_{u_{|C|}})$ denotes the sequence of representations of utterances.
For the $i$-th turn $u_i$, the embedding of $\mathtt{\langle bos \rangle}$ is chosen as its representation, i.e., $h_{u_i} = \mathbf{H}^{\mathcal{T},N}_{i,0}$. Different from the token encoder, we propose the Thread-Aware Attention sub-layer to replace the self-attention sub-layer to encode the tree-structure information into our model.

\paragraph{Thread-Aware Attention.} 
Each sub-layer consists of $h$ attention heads, and the results from each head are concatenated together and projected to form the output of the sub-layer. 
Formally, given the input $\mathbf{H}^{\mathcal{U},0}$, the $k$-th head computes a new sequence $z_k=(z_{k,0}, \cdots, z_{k,|C|})$ by:
\begin{equation}\small
    \label{eq:abs-attn}
    \begin{split}
        z_{k,i} = \sum_{j=1}^{|x|} \alpha_{ij}(h_{u_j}W_k^V), 
        \alpha_{ij} = \frac{\exp e_{ij}}{\sum^{|x|}_{t=1}\exp e_{it}}
    \end{split}
\end{equation}
where $z_{k,i} \in \mathbb{R}^{d_z}, d_z=d_h/h$. $e_{ij}$ is the attention weight from $h_{u_j}$ to $h_{u_i}$. Inspired by the relative position encoding (RPE) works \cite{shaw2018self,huang-etal-2020-improve}, we consider the interactions of queries, keys, and relative positions simultaneously to fully utilize the structural information of a conversation:
\begin{equation}\small
    \label{eq:relative-attn}
    e_{ij} = \frac{(h_{u_i}W_k^Q+r_{ij})(h_{u_j}W_k^K+r_{ij})^\top - r_{ij}r_{ij}^\top}{\sqrt{d_z}}
\end{equation}
where $W_k^Q,W_k^K,W_k^V \in \mathbb{R}^{d_h\times d_z}$ are parameter matrices. $\sqrt{d_z}$ is a scaling factor for stable training. The key to this mechanism is that $r_{i,j}\in \mathbb{R}^{d_z}$ encodes the relation from utterance $u_j$ to $u_i$, which is defined as:
\begin{equation}\small
    \label{eq:thread-aware-emb}
    r_{i,j} = 
    \begin{cases}
        w_{\mathrm{clip}(\mathrm{depth}(u_i)-\mathrm{depth}(u_j), k)}, &1)\\
        w_*, &2)
    \end{cases}
\end{equation}

\begin{figure}[t]
    \centering
    \includegraphics[width=\columnwidth]{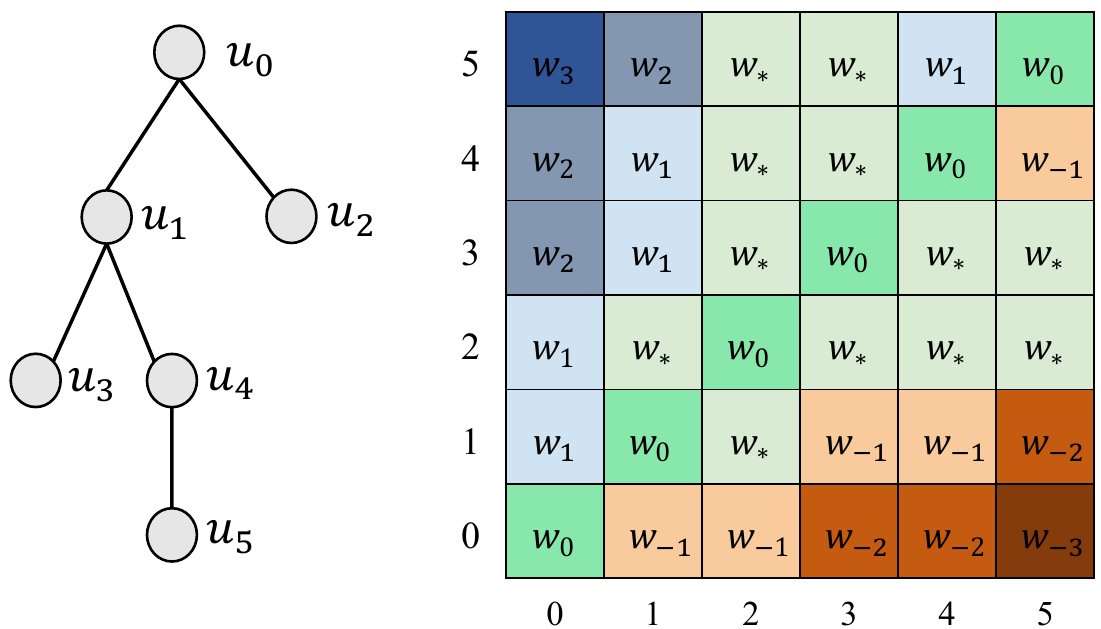}
    \caption{Illustration of Thread-Aware attention. The left is the tree structure of a conversation thread, and $u_i$ represents the $i$-th utterance. The thread-aware attention weights across the utterances are on the right.}
    \label{fig:rel-attn}
\end{figure}

As illustrated in Figure \ref{fig:rel-attn}, the relation between two utterances has two situations: 1) one is a parent or child utterance of the other, that is, they belong to the same path, e.g. $u_1$ and $u_4$; 2) otherwise, e.g. $u_2$ and $u_3$. We totally define $2k+2$ learnable thread-aware position embeddings $\{w_*, w_{-k}, \cdots, w_k\}$. 
where $\mathrm{clip}(x,k)=\max(-k,\min(k,x))$, and the function $\mathrm{depth}(u_i)$ returns the distance between utterance $u_i$ and the first utterance $u_0$ in the thread, e.g. $\mathrm{depth}(u_4)=2$. 
The output of the utterance encoder is $\mathbf{H}^{\mathcal{U}, N} \in \mathbb{R}^{|C|\times d_h}$.

\paragraph{Decoder.} The decoder is a $N$-layer transformer to generate the summary $S$. At the training stage, the decoder takes the right-shifted token sequence of $S$ as input. 
In each layer, the self-attention sub-layer leverages a lower triangular mask to prevent positions from attending to their future positions. Then, the cross-attention sub-layers attend with the outputs from the hierarchical encoder.
In particular, we make an encoder-wise residual connection around the utterance encoder to propagate the token-level information directly to the decoder. We found this can improve the model's capability to reproduce the words involved in the conversation. Denote the output of the decoder as $\mathbf{H}^{\mathcal{D},N} \in \mathbb{R}^{|S|\times d_h}$. When predicting the $i$-th token $s_i$, we reuse the embedding matrix of the vocabulary $\mathcal{E}_{\mathcal{V}} \in \mathbb{R}^{|\mathcal{V}|\times d_h}$ to project $\mathbf{H}^{\mathcal{D},N}_{i-1}$ into a probability distribution:
\begin{equation}\small
    \label{eq:decode-prob}
    P(S_i|S_{<i}, C) = \mathrm{Softmax}(\mathbf{H}^{\mathcal{D},N}_{i-1}\mathcal{E}_{\mathcal{V}}^\top)
\end{equation}

\subsection{Pretraining Objectives}
In this section, we describe the pretraining objectives used for pretraining \textsc{TANET}. In addition to the causal language modeling, we newly introduce another thread-aware pretraining task to predict the contextual relation between utterances.

\paragraph{Causal Language Modeling.} Following many previous works \cite{lewis-etal-2020-bart,zhang2020pegasus}, we apply the causal language modeling objective, which seeks to minimize the cross-entropy loss:
\begin{equation}\small
    \label{eq:clm}
    \mathcal{L}_{CLM}(\theta) = -\frac{1}{|S|}\sum^{|S|}_{i=1} \log P(S_i|S_{<i},C)
\end{equation}

\paragraph{Thread Prediction.} To enhance the representation of the thread structure in a conversation, we introduce a new pretraining task of thread prediction. The motivation is to encourage the model to learn thread-aware representations that encode the information of which comments this one was written based on, that is, its historical comments.
Specifically, we randomly sample $20\%$ utterances $C_s$ from $C$ and then let the model predict their historical comments. Formally, the pretraining objective is calculated as:
\begin{equation}\small
    \label{eq:thread-pred}
    \begin{split}
        \mathcal{L}_{ThreadPred} = -&\sum_{a_{ij}\in \mathcal{A}}\Big( \delta(a_{ij})\log p_{a_{ij}} \\
        &+ \big(1-\delta(a_{ij})\big)\log (1-p_{a_{ij}})\Big)  \\  
    \end{split}
\end{equation}
where $\mathcal{A}=C_s\times C \cup C\times C_s$ is the set of comment pair candidates for prediction. $\delta(a_{ij})$ returns $1$ if $u_j$ and $u_i$ belongs to one thread and $u_j$ is history of $u_i$, otherwise $0$. $p_{ij}$ is the probability of $u_j$ being the historical comment of $u_i$ and is computed by:
\begin{equation}\small
    \label{eq: history-prob}
    p_{a_{ij}} = \mathrm{Sigmoid}\big((\mathcal{H}^{\mathcal{T},N}_{i,0} W_a) ({\mathcal{H}^{\mathcal{T},N}_{j,0}}W_b)^\top \big)
\end{equation}
$W_a, W_b \in \mathbb{R}^{d_h\times d_h}$ are two parameter matrices.

\subsection{Application on Downstream Tasks}
After the pretraining stage, we finetune our model on the downstream tasks. Different downstream tasks will have some differences in structure that requires us to adapt it flexibly. For example, we cannot know the interdependence between the utterances in the meeting, so we regard it as a sequence of utterances in one thread. In addition, some additional information of the data is very necessary for the generation of the summary. As \citet{zhu-etal-2020-hierarchical} said, the role of the participant is very useful in meeting summary genration. Without specific design, we inform \textsc{TANET} of the information by modifying the utterance with template  \{\textit{participant}\} of role \{\textit{role}\} said: \{\textit{utterance}\}.
\section{Experiments}
\label{sec:exp}
\subsection{Datasets}
We evaluate \textsc{TANet} and all baseline models on four benchmark datasets of long and real-life conversations, covering domains of meeting transcripts, customer-service records, and threads in web forum. Table \ref{tab:stat} summarizes the statistics of the four datasets.

\noindent
\textbf{AMI} \cite{10.1007/11677482_3} is a multi-modal dataset consisting of 100 hours of meeting recordings with rich annotations.
Following \citet{shang-etal-2018-unsupervised,zhu-etal-2020-hierarchical}, we select $137$ meetings of scenario where the participants play different roles in a design team. Each meeting is labeled with transcripts produced by automatic speech recognition (ASR) and an abstractive summary written by a human annotator. Furthermore, each dialogue is also associated with additional information, including its speaker id with role, dialogue act. We use the same data split of $100/17/20$ as training/validation/test sets.

\noindent
\textbf{ICSI} \cite{1198793} is another widely-used meeting corpus consisting of about 70 hours of meeting audio recordings with orthographic transcription and other manual annotations. We follow the pre-processing pipeline from \citet{zhu-etal-2020-hierarchical} and split the training/validation/test sets of size $43/10/6$, respectively. Each meeting also contains a manually labeled abstractive summary and the associated role information for each participant.

\noindent
\textbf{MultiWOZ} \cite{yuan2019abstractive} is an abstractive dialog summarization dataset based on the MultiWOZ corpus \cite{budzianowski2018large,ramadan2018large,eric2019multiwoz,zang2020multiwoz}, which is a fully-labeled collection of human-human written conversations spanning over multiple domains and topics. The dataset is built on various customer-service records in the corpus, such as booking restaurants, hotels, taxis. We use the summary annotation provided by \citet{yuan2019abstractive}, and the same data split of $8438/1000/1000$ as training/validation/test sets.

\noindent
\textbf{FORUM} \cite{DBLP:conf/flairs/TarnpradabLH17} contains $700$ human-annotated forum threads. Each thread contains a human-annotated abstractive summary and multiple posts written by several different users.
These threads are collected from tripadvisor.com and ubuntuforums.org.
\citet{DBLP:conf/emnlp/BhatiaBM14} annotated $100$ threads from TripAdvisor with human-writtern summaries, and \citet{DBLP:conf/flairs/TarnpradabLH17} further extend the summary annotation with $600$ more threads.
In our experiments, we divide the dataset into $500/100/89$ examples for training/validation/test sets.

\begin{table}[t!]
  \centering
  \resizebox{\columnwidth}{!}{
    \begin{tabular}{lcccc}
      \toprule
      Dataset          & AMI     & ICSI    & MultiWOZ         & FORUM        \\
      \midrule
      Domain           & \textit{Meeting} & \textit{Meeting} & \textit{Customer Service} & \textit{Forum Thread} \\
      \# Speakers      & 4       & 6.2     & 2                & 6.8          \\
      \# Conversations & 137     & 59      & 10,438           & 689          \\
      \# Conv. words   & 4,757   & 10,189  & 180.7            & 825.0        \\
      \# Summ. words   & 322     & 534     & 91.9             & 190.6        \\
      \# Turns         & 289     & 464     & 13.7             & 10.5         \\
      \bottomrule
    \end{tabular}}
  \caption{Statistics of the converstional summarization datasets. The number of conversation words, summary words, turns and speakers are all averaged across all conversations in the dataset.}
  \label{tab:stat}
\end{table}

\subsection{Evaluation Results and Discussions}
\label{sec:experiments}
\subsubsection{Meeting Summarization}
We compare \textsc{TANet} with a variety of models from previous literature: \textbf{Random} \cite{DBLP:conf/interspeech/RiedhammerGFH08}, the template-based model \textbf{Template} \cite{DBLP:conf/inlg/OyaMCN14}, two ranking systems \textbf{TextRank} \cite{DBLP:conf/emnlp/MihalceaT04} and \textbf{ClusterRank} \cite{DBLP:conf/interspeech/GargFRH09}, the unsupervised method \textbf{UNS} \cite{shang-etal-2018-unsupervised}, \textbf{Extractive Oracle}, which concatenates top sentences with the highest ROUGE-1 scores with the golden summary, the document summarization model \textbf{PGNet} \cite{see-etal-2017-get}, \textbf{Copy from Train} which randomly copies a summary from the training set as the prediction, the multimodal model \textbf{MM} \cite{li-etal-2019-keep}, and the hierarchical Network \textbf{HMNet} \cite{zhu-etal-2020-hierarchical}. Besides the baselines above, \textbf{BART} \cite{lewis-etal-2020-bart} and \textbf{PEGASUS} \cite{zhang2020pegasus}, two state-of-the-art pre-trained models on document summarization, and \textbf{Longformer-Encoder-Decoder} (\textbf{LED}) \cite{DBLP:journals/corr/abs-2004-05150} are also included in comparison to have a thorough understanding towards our model. We concatenate all turns of a transcript into a sequence and then truncate it to meet the length constraints of the models' input.
LED$_{large}$ is initialized from BART$_{large}$ and able to process $16$k tokens.
Please refer to the Appendix for more implementation details .
\begin{table}[t!]
  \centering
  \resizebox{\columnwidth}{!}{
    \begin{tabular}{l ccc}
      \toprule
      Models                                   & R-1            & R-2            & R-SU4          \\  
      \midrule
      \multicolumn{4}{c}{AMI}                                                                    \\  
      \midrule
      Random \cite{DBLP:conf/interspeech/RiedhammerGFH08}                & 35.13                   & 6.26                     & 13.17                    \\
       Template \cite{DBLP:conf/inlg/OyaMCN14} & 31.50                   & 6.80                     & 11.40 \\
       TextRank \cite{DBLP:conf/emnlp/MihalceaT04}              & 35.25                   & 6.90                     & 13.62                    \\
       ClusterRank \cite{DBLP:conf/interspeech/GargFRH09}           & 35.14                   & 6.46                     & 13.35         \\
      UNS \cite{shang-etal-2018-unsupervised}                   & 37.86                   & 7.84                     & 14.71\\
       Extractive Oracle      & 39.49                   & 9.65                     & 13.20       \\
      PGNet \cite{see-etal-2017-get}          & 40.77          & 14.87          & 18.68          \\
      Copy from Train                                  & 43.24                   & 12.15                    & 14.01                    \\
      MM+TopicSeg \cite{li-etal-2019-keep}$^\dagger$     & 51.53                   & 12.23                    & -              \\
      MM+TopicSeg+VFOA \cite{li-etal-2019-keep}$^\dagger$                    & \textbf{53.29} & 13.51          & -              \\
    HMNet \cite{zhu-etal-2020-hierarchical} & 53.02          & 18.57          & 24.85          \\
      \midrule
      \textit{Our re-implementation} & & &\\
      LED$_{large}$ \cite{DBLP:journals/corr/abs-2004-05150} & 53.10& 19.83& 24.95\\
      BART$_{base}$  \cite{lewis-etal-2020-bart}  & 50.26          & 18.18          & 17.83          \\
      PEGASUS$_{large}$  \cite{zhang2020pegasus}  & 47.05          & 16.64          & 16.03          \\
      \midrule
    \textsc{TANet} (\textit{ours})            & 53.26          & \textbf{20.73}$^*$ & \textbf{25.98}$^*$ \\
    \toprule

      \multicolumn{4}{c}{ICSI}                                                                   \\
    \midrule
       Random \cite{DBLP:conf/interspeech/RiedhammerGFH08}           & 29.28          & 3.78           & 10.29          \\
       TextRank \cite{DBLP:conf/emnlp/MihalceaT04}  & 29.70           & 4.09           & 10.64          \\
       ClusterRank \cite{DBLP:conf/interspeech/GargFRH09}         & 27.64          & 3.68           & 9.77           \\
       UNS \cite{shang-etal-2018-unsupervised} & 31.60          & 4.83           & 11.35          \\
       Extractive Oracle         & 34.66          & 8.00           & 10.49          \\
      PGNet \cite{see-etal-2017-get}          & 32.00          & 7.70           & 12.46          \\
       Copy from Train  & 34.65          & 5.55           & 10.65          \\
      HMNet \cite{zhu-etal-2020-hierarchical} & 46.28          & 10.60          & 19.12          \\
      \midrule
    \textit{Our re-implementation} & & &\\
      LED$_{large}$ \cite{DBLP:journals/corr/abs-2004-05150} & 43.13& 11.76& 19.08\\
      BART$_{base}$  \cite{lewis-etal-2020-bart}  & 42.01          & 9.96           & 11.72          \\
      PEGASUS$_{large}$  \cite{zhang2020pegasus}  & 42.44          & 9.15           & 11.10          \\
      \midrule
     \textsc{TANet} (\textit{ours})            & \textbf{47.21}$^*$ & \textbf{12.35}$^*$ & \textbf{19.27} \\
      \bottomrule
    \end{tabular}
  }
  \caption{Automatic evaluation results on datasets AMI and ICSI.
    Numbers in bold indicate the best performing models on the corresponding metrics. 
    Numbers marked with ``*'' mean that the improvement over the best baseline is statistically significant (t-test with $p$-value $<0.05$).
    Models marked with ``$\dagger$'' require additional human annotations of topic segmentation and visual signals from cameras.
  }
  \label{tab:meeting_results}
\end{table}

Table \ref{tab:meeting_results} reports the evaluation results of ROUGE metrics \cite{lin2004rouge} on datasts AMI and ICSI.
We can see that, except for ROUGE-1 on AMI, \textsc{TANet} outperforms all baseline models in all metrics. 
MM is a multi-modal model which requires additional annotation of topic segmentation (TopicSeg) and multi-modal features derived from the visual focus of attention (VFOA) collected by cameras. In practice, the visual information is rarely available, such as online chat, so the application scenarios of MM are very limited. In comparison, \textsc{TANet} is completely based on meeting transcripts from ASR systems, so it has better scalability. Comparable performance is achieved in ROUGE-1 on AMI, but it is significantly higher in ROUGE-2 by $7.2$ points. 
In particular, \textsc{TANet} outperforms HMNet, indicating that pretraining on large-scale conversation data while incorporating the inherent structural information can lead to better performances on downstream tasks. 
Moreover, \textsc{TANet} significantly outperforms BART and PEGASUS on both AMI and ICSI. 
Although the two baselines own strong capabilities to summary a document, the tricky part is that a meeting transcript is very long and cannot be fully fed into the models. For example, the average number of words in ICSI is $10,189$, which far exceeds the maximum input length $512$ tokens of BART and PEGASUS. As a result, most of the content in a meeting transcript are discarded, which will inevitably limit the performances of the two models. 
LED can input all sentences, but it is still difficult to fully understand the conversation which further demonstrate the effectiveness of the pretraining on the corpus RCS.

\begin{table}[t!]
  \centering
  \resizebox{\columnwidth}{!}{
    \begin{tabular}{l ccc}
      \toprule
      Models                                   & R-1            & R-2            & R-L            \\
      \midrule
      PGNet \cite{see-etal-2017-get}          & 62.89          & 48.61          & 59.30          \\
      Transformer \cite{vaswani2017attention} & 63.12 & 50.63& 61.04 \\
      SPNet \cite{yuan2019abstractive}        & 90.97          & 84.14          & 85.00          \\
      \midrule
      \textit{Our re-implementation} & & &\\
      LED$_{large}$ \cite{DBLP:journals/corr/abs-2004-05150} & 91.41& 79.93& 83.63\\
      HMNet \cite{zhu-etal-2020-hierarchical} & 66.33          & 50.49          & 64.52          \\
      BART$_{base}$  \cite{lewis-etal-2020-bart}  & 81.47          & 70.24          & 73.14          \\
      PEGASUS$_{large}$  \cite{zhang2020pegasus}  & \textbf{93.51} & 88.09 & 84.73          \\
      \midrule
      \textsc{TANet} (\textit{ours})            & 93.25          & \textbf{88.60}$^*$          & \textbf{85.67}$^*$ \\
      \bottomrule
    \end{tabular}
  }
  \caption{Automatic evaluation results on MultiWOZ. Numbers in bold indicate the best performing models on the corresponding metrics.
  Numbers marked with ``*'' mean that the improvement over the best baseline is statistically significant (t-test with $p$-value $<0.05$).
  }
  \label{tab:csr_results}
\end{table}

\subsubsection{Customer-service Records Summarization}
To demonstrate the effectiveness of \textsc{TANet} on customer-service records summarization, following models are selected as baselines from previous literature: the pointer-generator network \textbf{PGNet} \cite{see-etal-2017-get}, \textbf{Transformer} \cite{vaswani2017attention} and \textbf{SPNet} which incorporates three types of semantic scaffolds - speaker role, semantic slot and dialogue domain for summarization \cite{yuan2019abstractive}. Besides, We include \textbf{HMNet} \cite{zhu-etal-2020-hierarchical} as a baseline and implement it using the official code \footnote{\url{https://github.com/microsoft/HMNet}}.
We also apply \textbf{Longformer-Encoder-Decoder} (\textbf{LED}) \cite{DBLP:journals/corr/abs-2004-05150}, \textbf{BART} \cite{lewis-etal-2020-bart} and \textbf{PEGASUS} \cite{zhang2020pegasus} in this task by concatenating all utterances in a conversation as a document. 

Table \ref{tab:csr_results} reports the evaluation results of ROUGE metrics \cite{lin2004rouge} on MultiWOZ.
We can observe that \textsc{TANet} achieves new state-of-the-art performance on ROUGE-2 and ROUGE-L, which demonstrate the effectiveness of pretraining on large-scale conversation data. 
Different from the results on meeting summarization given in Table \ref{tab:meeting_results}, PEGASUS$_{large}$ achieves close performance to \textsc{TANet} and even the best in ROUGE-1, showing its great generalization ability in this task.
This is because: (1) the ``documents'' can be fully fed into the model without content loss. 
The average length of the dialogue in MultiWOZ is $180.7$ words, which do not exceed the model's maximum input length $512$. (2) each conversation takes place between two speakers (i.e. a customer and a staff), so the structure of a dialogue can be viewed as a sequence, which is similar to the sentences in a document. 
Compared with LED, HMNet and BART, the un-pretrained model SPNet obtains surprisingly better scores. This motivates us to combine richer conversation-related information, such as speaker role, dialogue act, semantic slot, and dialogue domain, to further improve model's summarization capabilities in the future.

\begin{table}[t!]
  \centering
  \resizebox{\columnwidth}{!}{
    \begin{tabular}{lccc}
      \toprule
      Models                                            & R-1   & R-2   & R-L            \\
      \midrule
      ILP \cite{DBLP:conf/acl/Berg-KirkpatrickGK11}                                              & 29.3  & 9.9   & -              \\
      Sum-Basic \cite{DBLP:journals/ipm/VanderwendeSBN07}                                        & 33.1  & 10.4  & -              \\
      KL-Sum                                           & 35.5  & 12.3  & -              \\
      Lex-Rank \cite{DBLP:journals/corr/abs-1109-2128}                                         & 38.7  & 14.2  & -              \\
      MEAD \cite{DBLP:journals/ipm/RadevJST04}                                             & 38.5  & 15.4  & -              \\
      SVM \cite{DBLP:journals/tist/ChangL11}                                              & 24.7  & 10.0  & -              \\
      LogReg \cite{DBLP:journals/jmlr/FanCHWL08}                                     & 29.4  & 7.8   & -              \\
      HAN \cite{DBLP:conf/flairs/TarnpradabLH17}  & 37.8  & 14.7  & -              \\
    \midrule
    \textit{Our re-implementation} & & &\\
    LED$_{large}$ \cite{DBLP:journals/corr/abs-2004-05150} & 42.39& 22.78& 30.48\\
    HMNet \cite{zhu-etal-2020-hierarchical}          & 41.30 & 17.12 & 31.76 \\
      BART$_{base}$ \cite{lewis-etal-2020-bart}                 & 42.91 & 22.32 & 30.35          \\
      PEGASUS$_{large}$ \cite{zhang2020pegasus}                  & 42.92 & 20.50 & 29.16          \\
      \midrule
      \textsc{TANet} (\textit{ours})                                   & \textbf{45.20}$^*$ & \textbf{25.61}$^*$ & \textbf{33.59}$^*$          \\
      \bottomrule
    \end{tabular}
  }
  \caption{Automatic evaluation results on FORUM. Numbers in bold indicate the best performing models on the corresponding metrics.
  Numbers marked with ``*'' mean that the improvement over the best baseline is statistically significant (t-test with $p$-value $<0.05$).}
  \label{tab:forum_results}
\end{table}

\subsubsection{Forum Threads Summarization}
In this task, \textsc{TANet} is compared against a range of baselines, including following unsupervised methods: 1) \textbf{ILP} \cite{DBLP:conf/acl/Berg-KirkpatrickGK11}, a baseline integer linear programming framework; 2) \textbf{Sum-Basic} \cite{DBLP:journals/ipm/VanderwendeSBN07}, a model that assumes words occurring frequently in a document cluster have a higher chance of being included in the summary; 3) \textbf{KL-Sum}, an approach that select the sentences decreasing the KL divergence as the summary; 4) \textbf{Lex-Rank} \cite{DBLP:journals/corr/abs-1109-2128}, a graph-based model based on eigenvector centrality; 5) \textbf{MEAD} \cite{DBLP:journals/ipm/RadevJST04}, a centroid-based approach which scores sentences based on length, centroid, and position; 
and supervised extractive systems, including 1) \textbf{SVM} \cite{DBLP:journals/tist/ChangL11}, the support vector machine; 2) \textbf{LogReg} \cite{DBLP:journals/jmlr/FanCHWL08}, the logistic regression; 3) \textbf{HAN} \cite{DBLP:conf/flairs/TarnpradabLH17}, a hierarchical attention network with redundancy removal process;
Besides, we also apply the cross-domain pre-trained model \textbf{HMNet} \cite{zhu-etal-2020-hierarchical} in this task and implement \textbf{Longformer-Encoder-Decoder} (\textbf{LED}) \cite{DBLP:journals/corr/abs-2004-05150}, \textbf{BART} \cite{lewis-etal-2020-bart} and \textbf{PEGASUS} \cite{zhang2020pegasus} in a similar way to the adaptation in the above two tasks.

Table \ref{tab:forum_results} reports the evaluation results of ROUGE metrics \cite{lin2004rouge} on the dataset FORUM. \textsc{TANet} outperforms all baseline models in terms of all metrics, and the improvements are statistically significant (t-test with $p$-value$<0.05$), which further demonstrate the effectiveness of our method. 
In this task, the gains over the pre-trained baselines are relatively high , due to (1) the consistency of conversation domain between the pretraining stage and downstream fine-tuning. The conversation in FORUM and our pretraining corpus RCS are both forum threads. Note that, although the data in FORUM is collected from TripAdvisor (tripadvisor.com) and UbuntuForums (ubuntuforums.org.), the subjects are also included in some specific sub-reddits on the Reddit website. In contrast, LED, HMNet, BART and PEGASUS are all pre-trained with document-like text, so there will be a domain gap in threads' understanding;
(2) structure modeling. The tree-like reply relationship in a thread plays a vital role in understanding the entire thread, but the baselines can only process it linearly, which poses challenges for the model to fully understand the context and generate accurate summaries. 

\begin{table}[t!]
  \centering
  \resizebox{\columnwidth}{!}{
    \begin{tabular}{l|ccc|ccc}
      \thickhline
      \multicolumn{1}{c|}{\multirow{2}{*}{Models}} & 
      \multicolumn{3}{c|}{AMI}   &\multicolumn{3}{c}{FORUM} \\
      
      \cline{2-7}
                                          & R-1  & R-2  & R-SU4  & R-1  & R-2  & R-L\\
      \hline
      \textsc{TANet}                      & 53.26 & 20.73 & 25.98  & 45.20 & 25.61 & 33.59\\
      - \textit{Pretraining}                    & 46.43 & 16.85 & 18.42  & 39.67 & 15.37 & 26.45\\
      - \textit{Thread-Aware Attention}          & 51.33 & 18.90 & 23.71  & 41.85 & 22.41 & 30.79\\
      - \textit{Thread Prediction}               & 51.94 & 19.30 & 24.75  & 41.70 & 21.33 & 31.80\\
      - \textit{Encoder-wise Residual}           & 51.86 & 20.02 & 24.59  & 44.93 & 24.78 & 33.30\\
      \thickhline
    \end{tabular}
  }
  \caption{Ablation study on AMI and FORUM.}
  \label{tab:ablation}
\end{table}

\subsubsection{Ablation Study}
To understand the impact of our pretraining strategies on model performance, we compare the full \textsc{TANet} with the following variants: (1) -\textit{Pretraining}: the pretraining stage is removed; (2) - \textit{Thread-Aware Attention}: the Thread-Aware Attention sublayers in the utterance encoder degenerate into standard self-attention sublayers; (3) -\textit{Thread Prediction}: $\mathcal{L}_{ThreadPred}$ is removed; and (4) -\textit{Encoder-wise Residual}: the encoder-wise residual around the utterance encoder is removed. Table \ref{tab:ablation} reports the evaluation results on AMI and FORUM.\footnote{Ablation results on AMI and FORUM could provide more insights, whereas ICSI is similar to AMI, and MultiWOZ is less challenging.} We can conclude that (1) the pretraining on RCS helps to significantly improve the performance, as removing it results in dramatic performance drop on both AMI and FORUM; (2) both Thread-Aware attention and Thread Prediction objective are useful, indicating that the structure of thread is essential to facilitate the understanding of conversation, especially for the threads with tree structure; (3) the encoder-wise residual is meaningful, as removing it causes performance drop.

\subsubsection{Human Evaluation} 
As the human annotation for this task is very time-consuming and labor-intensive, we also conduct human evaluation on the test sets of AMI and FORUM to verify whether the improvements on automatic evaluation is in line with the human perceived quality. 
We recruit $3$ well-educated native speakers as annotators and compare \textsc{TANet} with BART$_{base}$ and HMNet on $2$ aspects - \textit{readability} and \textit{conciseness}. The former measures how fluent a generated summary is, while the later measures how well the summary sums up the main ideas of a conversation.
For each sample, we show its conversation, reference summary, as well as summaries generated by models (the order is shuffled to hide their sources) to the annotators and ask them to judge the quality and assign a score in \{0,1,2\} (indicating ``bad'', ``fair'', and ``good'') to each summary for each aspect. 
Table \ref{tab:human-eval} reports the evaluation results. We can observe (1) the three models are comparable on readability on both datasets; (2) \textsc{TANet} outperforms the others on conciseness, which is consistent with the automatic evaluation results in Table \ref{tab:meeting_results} and Table \ref{tab:forum_results}; (3) Bart$_{base}$ does not perform well in conciseness on AMI, as most utterances of a conversation are discarded due to the input length constraint $512$.  All kappa values are no less than 0.6, indicating substantial agreement among the annotators. For reference, we present case study in the Appendix.

\begin{table}[t!]
  \centering
  \resizebox{0.9\columnwidth}{!}{
    \begin{tabular}{l|ccc|ccc}
      \thickhline
      \multicolumn{1}{c|}{\multirow{2}{*}{Models}} & 
      \multicolumn{3}{c|}{AMI}   &\multicolumn{3}{c}{FORUM} \\
      
      \cline{2-7}
                                          & Read.  & Conc.  & Kappa  & Read.  & Conc.  & Kappa\\
      \hline
      HMNet                               & 1.67 & 1.40 & 0.60  & 1.76 & 1.57 & 0.61\\
      Bart$_{base}$                       & 1.58 & 1.02 & 0.72  & 1.82 & 1.63 & 0.69\\
      \textsc{TANet}                      & 1.70 & 1.53 & 0.63  & 1.84 & 1.70 & 0.60\\
      \thickhline
    \end{tabular}
  }
  \caption{Human evaluation results on AMI and FORUM. ``Read.'', ``Conc.'' are abbreviations for readability and conciseness, respectively.}
  \label{tab:human-eval}
  \vspace{-2mm}
\end{table}
\section{Related Work}
With the recent success of seq2seq models, the research focus of conversational summarization has been transferred from the extractive methods to abstractive models. Various semantic patterns have been applied to these abstractive approaches, such as dialogue acts \cite{8639531}, auxiliary key point sequences \cite{10.1145/3292500.3330683}, topic segments \cite{li-etal-2019-keep,9003764}, conversational stages and dialogue overview \cite{chen-yang-2020-multi}, discourse relations \cite{murray-etal-2006-incorporating,bui-etal-2009-extracting,qin-etal-2017-joint}. At the same time, some work is devoted to providing high-quality datasets to promote the development of this research direction \cite{10.1007/11677482_3,1198793,DBLP:conf/flairs/TarnpradabLH17,yuan2019abstractive,gliwa-etal-2019-samsum}. However, these corpora have a low number of conversations which hinders the progress of abstractive summarization \cite{DBLP:journals/corr/abs-2107-03175}. Recently, large neural models pre-trained on huge corpora have led to strong improvements on numerous natural language understanding and generation tasks \cite{devlin-etal-2019-bert,radford2019language,lewis-etal-2020-bart,zhang2020pegasus}. Encouraging by the promising progress of pretraining, \citet{zhu-etal-2020-hierarchical} first introduce a hierarchical structure and propose pretraining on cross-domain data for meeting summarization. The pretraining data is collected from the news domain. Regarding one document as the utterances from one participant, multiple documents are combined and reshuffled to simulate a multi-person meeting. However, there are two disadvantages - the first is the style inconsistency between conversation and news, and the second is that there is no contextual relationship between the two documents, so the participants have no communication actually.
In this work, we build a large-scale corpus based on real conversations. Besides, we further incorporate the structure information of thread in the model.
\section{Conclusions}
\label{sec:conclusion}
In this work, we introduce \textsc{TANET}, a thread-aware pre-trained model for abstractive conversational summarization. 
\textsc{TANET} employ the thread-aware attention and a new pretraining objective to fully leverage the structure information of conversation. 
Furthermore, we build a large-scale pretraining corpus based on the discussions on Reddit. Experiments on four downstream tasks demonstrate the effectiveness of \textsc{TANET}. 


\bibliography{custom}
\bibliographystyle{acl_natbib}

\appendix

\section{Implementation Details}
\label{sec:appendix}
In \textsc{TANet}, the token encoder, utterance encoder, and decoder all have $6$ layers, i.e., $N=6$. Each multi-head attention sub-layer has $12$ heads, i.e., $h=12$. The size of feed-forward layer is $3072$. The hidden size $d_h$ is $768$.
We employ the same vocabulary as BART \cite{lewis-etal-2020-bart}, which has $50265$ tokens. \textsc{TANet} has $180$M parameters in total. We use a dropout probability of $0.1$ for all layers. In the thread-aware attention layer, we define $20$ learnable embeddings, i.e., $k=9$.
For optimization, both pretraining and downstream finetuning use AdamW optimizer \cite{loshchilov2017decoupled} with $\beta_1=0.9$, $\beta_2=0.999$, and $\epsilon=1e-8$. We pre-train \textsc{TANet} with a accumulated batch size of $256$.
The initial learning rate is set as $5e-5$ and linearly decreased to $0$ after $500$k steps.
We use beam search with the commonly used trigram blocking \cite{paulus2017deep,lewis-etal-2020-bart} to select the best candidate during inference for the downstream tasks.
To improve the pretraining efficiency, we set the maximum number of utterances to $124$, each utterance has a maximum of $200$ tokens, and the pseudo-summary has a maximum of $256$ tokens. 
BART$_{base}$ and PEGASUS$_{large}$ are implemented with the codes provided by HuggingFace at \url{https://github.com/huggingface/transformers/tree/v4.1.1/examples/seq2seq}. The initialized pre-trained models are available at \url{https://huggingface.co/facebook/bart-base} and \url{https://huggingface.co/google/pegasus-large}. We implement LED$_{large}$ by referring to the code at \url{https://colab.research.google.com/github/patrickvonplaten/notebooks/blob/master/Fine_tune_Longformer_Encoder_Decoder_(LED)_for_Summarization_on_pubmed.ipynb}.
All experiments are conducted on a machine with $4$ NVIDIA Tesla V100  GPUs.

\section{Case Study}
Finally, we randomly sample two cases from the test sets of MultiWOZ and FORUM to conduct qualitative analysis. 
We show the conversation and the summaries generated by HMNet, LED$_{large}$, Bart$_{base}$, \textsc{TANet}, as well as the ground truth in Table \ref{tab:multiwoz-c1} and Table \ref{tab:forum-c1}. We can see that \textsc{TANet} can produce a more concise summary with less incorrect facts and has a higher level of abstraction ability about the conversations. In contrast, the summaries generated by HMNet and Bart$_{base}$ cannot cover all the important information and are prone to some mistakes. LED$_{large}$ tends to copy content from the original conversation, especially as in Table \ref{tab:forum-c1}.

\begin{table*}[th]
    \small
    \centering
    \resizebox{\linewidth}{!}{
        \begin{tabular}{p{14mm}p{14cm}  }
            \hline
            \multicolumn{2}{c}{\textbf{Conversation}}                                                                                                                                                                                                                                                                                                                                                                                                                                                                                                                                                                                                                                             \\
            \hline
            \textbf{USER}:         & I'm \underline{looking for information on Cambridge University Botanic Gardens}.                                                                                                                                                                                                                                                                                                                                                                                                                                                                                                                                                                              \\
            \textbf{SYSTEM}:      & They are on bateman st, postal code cb21jf, they can be reach at 01223336265, the entrance fee is 4 pounds, can I help with anything else?                                                                                                                                                                                                                                                                                                                                                                                                                                                                                                                  \\
            \textbf{USER}:         & Yes can you help me \underline{find a restaurant}?                                                                                                                                                                                                                                                                                                                                                                                                                                                                                 \\
            \textbf{SYSTEM}:      & The Botanic Gardens are in the centre. Would you like the restaurant to also be in the center? Do you have any type of cuisine in mind?                                                                                                                                                                                                                                                                                                                                                                                                                                                                                                                                               \\
            \textbf{USER}:         & Never mind, I will worry about food later. \underline{I am actually looking for a hotel with a guesthouse and free} \underline{parking would be great as well}.                                                                                                                                                                                                                                                                                                                                                                                                                                                                                                                                                                                                                     \\
            \textbf{SYSTEM}:      & There are 21 guesthouses with free parking, do you have a price or area preference?                                                                                                                                                                                                                                                                                                                                                                                                                                                                                                \\
            \textbf{USER}:         & \underline{\textcolor{blue}{ Cheap }} and in the south please.                                                                                                                                                                                                                                                                                                                                                                                                                                                                                    \\
            \textbf{SYSTEM}:      & I found a guest house listing for Rosa's Bed and Breakfast in the south. Would you like more information?                                                                                                                                                                                                                                                                                                                                              \\
            \textbf{USER}:         & \underline{I really need a hotel in the west please}.                                                                                                                                                                                                                                                                                                                                                                                                                                                                                                         \\
            \textbf{SYSTEM}:      & i have 1 hotel in the west. the cambridge belfry is cheap with 4 stars. it offers free parking and wifi. do you need a room?                                                                                                                                                                                                                                                                                                                                                                                                \\
            \textbf{USER}:         & I'm sorry.\underline{I really was looking for a guesthouse}.                                                                                                                                                                                                                                                                                                                                                                                                                                                                                                                                                                                                               \\
            \textbf{SYSTEM}:      & Okay. And did you want it in the west or the south?                                                                                                                                                                                                                                                                                                                                                                                                                                                                                                                                                                                          \\  \textbf{USER}: & I would like the west please. \\
            \textbf{SYSTEM}:      & I found finches bed and breakfast matching your request. Would you like to book a room now?  \\  
            \textbf{USER}: & Can you please \underline{book for Saturday, 2 people, 5 nights} please? \\
            \textbf{SYSTEM}:      & Booking was successful. \underline{Reference number is : W1QIVTNY} . \\  
            \textbf{USER}: & Thank you so much. I believe that is all I need today. Thanks again for all your help. \\
            \textbf{SYSTEM}:      & Have a nice day.                                                                                                                                                                                                                                                                                                                                                                                                                                                                                                                                                                                                             \\
            \\
            \hline
            \multicolumn{2}{c}{\textbf{Summaries}}                                                                                                                                                                                                                                                                                                                                                                                                                                                                                                                                                                                                                                                \\
            \hline
            \textbf{HMNet}:          &
            \textcolor{red}{you are traveling to and excited about seeing local tourist attractions} . you are looking for a place to stay . the hotel should be in the type of guest house and should include free parking . the restaurant should serve [ att\_name  ] food .\textcolor{red}{ make sure you get price range , area , and star of the hotel }. \\
            \\
            \textbf{LED$_{large}$}: &
            you are traveling to cambridge and looking forward to try local restaurants. you are looking for a particular attraction. its name is called [ att\_name ]. make sure you \textcolor{red}{get address}. you you are also looking for information in cambridge. you're looking for places to go in town. \textcolor{red}{the attraction should be in the type of museum. make sure to get postcode}. you also are looking to book a hotel. the hotel should be a guest house.
            \\
            \\
            \textbf{Bart$_{base}$}:  &
            \textcolor{red}{you are traveling to cambridge and excited about seeing local tourist attractions}. you are looking for a particular attraction. its name is called  [ att\_name ] . make sure you \textcolor{red}{get address}. you are also looking for a place to stay. the hotel should include free parking and should be in the type of guest house. the hotel \textcolor{red}{should be in the south} and should be in the cheap price range. once you find the hotel you want to book it for  [ people\_num ]  and  [ hotel\_stay ]  starting from  [ week\_day ] . make sure you get the reference number. \\
            \\
            \textbf{\textsc{TANet}}: &
           you are looking for information in cambridge. you are looking for a particular attraction. its name is called  [ att\_name ] . make sure you get \textcolor{red}{area} and phone number. you are also looking for a place to stay. the hotel should include free parking and should be in the type of guest house. the hotel should be in the \textcolor{blue}{cheap price} range. if there is no such hotel, how about one that is in the west. once you find the hotel you want to book it for  [ people\_num ]  and  [ hotel\_stay ]  starting from  [ week\_day ] . make sure you get the reference number. \\
            \\
            \textbf{Ground Truth}:   & you are looking for information in cambridge. you are looking for a particular attraction . its name is called  [ att\_name ] . make sure you get phone number and entrance fee. you are also looking for a place to stay . the hotel should be in the type of guest house and should include free parking. the hotel should be in the west. once you find the hotel you want to book it for  [ people\_num ]  and  [ hotel\_stay ]  starting from  [ week\_day ] . if the booking fails how about  [ hotel\_stay ] . make sure you get the reference number. \\
            \\
            \hline
        \end{tabular}
        }
    \caption{A case from MultiWOZ. 
    We underline some vital facts in the conversation. \textcolor{red}{Red} denotes the incorrect content in the generated summaries. \textcolor{blue}{Blue} indicates what appears in \textsc{TANet}'s summary but is not covered by the ground truth.}
    \label{tab:multiwoz-c1}
\end{table*}

\begin{table*}[t]
    \small
    \centering
    \resizebox{1\linewidth}{!}{
        \begin{tabular}{p{14.5mm}p{15cm}}
            \hline
            \multicolumn{2}{c}{\textbf{Conversation}}                                                                                                                                                                                                                                                                                                                                                                                                                                                                                                                                                                                                                                             \\
            \hline
            \textbf{N16E}:         & Hi, I'm hoping a local expert can help us out, we're \underline{traveling over to New York} (on route to Florida) on Wednesday 28th March, flying out on Saturday 31st, \underline{this gives us around 2 and a half days to see the city}, below is the list of places we are looking to visit/see. My only thoughts at the moment are to \underline{go to the ESB first thing around 8am and TOTR around dusk}. I was hoping to use the subway to get around and our hotel is The Belvedere just off 8th Ave on 48th street, in which order should we visit these sights? is it possible? any info on which subway lines to take would be fantastic. \underline{\textcolor{blue}{We are two families of 4} - 4 adults 4 kids}, aged 7 to 13. Maceys - browse for say 2 hours 5th Ave - stroll down and people watch \underline{Brooklyn Bridge} - wander over, check out the skyline \underline{Central Park} - relax \underline{Top of the Rock} - watch the transition from day to night \underline{Ground Zero} - must go and pay respect. \underline{Times Square} - sense the hustle and bustle Staton Island Ferry - relax a little \underline{Statue of Liberty} - view from the ferry? \underline{Empire State Building} - must do! \underline{Grand Central Station} - pass through and see the architecture \underline{Ellis Island} - not sure about this? \underline{Carnegie Deli} - take in a cheesecake. Have we missed anything? Given we are a party of 8 will we \underline{need to book restaurants? any suggestions nearby the hotel offering good steaks and pizza} (sorry I know this is a very subjective question). Thanks very much in advance - counting down the days. N16E                                                                                                                                                                                                                                                                                                                                                                                                                                                                                                                                                                            \\
            \textbf{SummerSh...}:      & Wow, that's a lot to try and squeeze into such a short visit! I'd eliminate Ellis Island and Macy's. Ellis Island just takes up too much time, and Macy's -- even though it's the original and the \"world's largest store\" -- is just a dept. store, you can find a branch in Florida too. \underline{You shouldn't have to make dinner reservations if you avoid the pre-theater crowd in the area.} Just go at 8:00. \underline{The best pizza is at John's, on W. 44th} (four blocks from your hotel). It's in a converted church, so very interesting. \underline{I will advise you to make reservations NOW for Carmine's}. Even though the place is enormous, it's really tough to get a reservation. That's where I'd suggest you get your steak fix. Their Flintstone-sized Porterhouse is yummy! \underline{Everything in Carmine's is served family style, so you'll only need two steaks for your entire group}. And one big bowl of pasta will feed everyone. If you go to the World Trade Center site, \underline{please also visit St. Paul's Chapel}, it's just a block away: www.saintpaulschapel.org And don't go to the Carnegie JUST for the cheesecake (even though it's fantastic!). You've gotta try their pastrami!                                                                                                                                                                                                                                                                                                                                                                                                                                                                                                                 \\
            \textbf{N16E}:         & Thanks Bettina, What order do you think we should visit these places in? Took a look at Carmines - the pictures on the website make it look a little formal? Thanks again. N16E.                                                                                \\
            \\
            \hline
            \multicolumn{2}{c}{\textbf{Summaries}}                                                                                                                                                                                                                                                                                                                                                                                                                                                                                                                                                                                                                                                \\
            \hline
            \textbf{HMNet}:          &
            \textcolor{red}{traveling over to new york on the route to florida on wednesday 28th march}. he is hoping to use the subway to take the belvedere . they are two families of 4 -4 adults and \textcolor{red}{they should visit around 8 am} . is also looking for advice on what to do on which subway lines to take and their hotel is fantastic . mentions that they are a party of 8 in advance and they will need to book down ellis island . they also recommend going to carnegie deli and \textcolor{red}{take pizza in carnegie deli} . recommends to \textcolor{red}{make reservations for 44th}.                                                                                                                                                                                                                                                                                                                        \\
            \\
            \textbf{LED$_{large}$}: &
            N16E is traveling to New York on Wednesday 28th March, flying out on Saturday 31st, gives them around 2 and a half days to see the city. They are two families of 4 - 4 adults 4 kids, aged 7 to 13. They have a list of places they are looking to visit/see. They would like to visit ESB first thing around 8am and TOTR around dusk. They were hoping to use the subway to get around and their hotel is The Belvedere just off 8th Ave on 48th street. They ask which order should they visit these sights? In which order? Any info on which subway lines to take would be fantastic. They also ask about steaks and pizza nearby the hotel offering good steaks. SummerShowers says to eliminate Ellis Island and Macy's. Ellis Island just takes up too much time, and Macy’s – even though it’s the original.
            \\
            \\
            \textbf{Bart$_{base}$}:  &
          N16E needed help to plan for a two day trip to New York. \textcolor{red}{N16E was traveling with his family of four} and had planned to visit the Empire State Building, Central Park, Top of the Rock, Times Square, Statue of Liberty, Grand Central Station, Ellis Island, Carnegie Deli, St. \textcolor{red}{Paul's Chapel, Macy's, and John's all in the same short time frame}. He/She also wanted to know the order of things to do in the city, and how to get around. Bettina said that they should make reservations NOW for Carmine`s. \\
            \\
            \textbf{\textsc{TANet}}: &
         N16E will be traveling to New York and needs help to plan for a two day stopover. They will be staying at the Belvedere Hotel, and would like to visit the Top of the Rock, and the Empire State Building. \textcolor{blue}{They are two families of four} and will be going to Macy`s. They plan on going to Ground Zero, Central Park, Times Square, Statue of Libery, and Grand Central Station. They also ask about dinner reservations, and suggestions on places that offer good steak and pizza.  SummerShowers mentions that  \textcolor{red}{John`s has a lot of steak and may be a little too formal}. He/She suggests Ellis Island, and to avoid the pre-theater crowd. Paul`s and Carmine`s are two family of four. N16E asks about a cheesecake in the Carnegie Deli. Summershowers responds that Carmine's is a good option as well and recommended a place called Le Pain de St. Paul's. For steak Carmine, go to Paul's Chapel and get pastrami for cheesecake. \\
            \\
            \textbf{Ground Truth}:   & N16E needed help to plan for a two day stopover in NYC en route to Florida, and wanted opinion on his/her itinerary. N16E said that they were staying at the Belvedere, and wanted to start at the Empire State Building and end at the Top of the Rock. N16E said that they had a party of 8 and that the Empire State Building was a must stop. N16E planned to browse at Macy's, stroll down the Brooklyn Bridge, go to Central Park to relax, visit the Top of the Rock, and see Ground Zero on the first day. On the second day N16E said that they would go to Times Square, ride the Staten Island Ferry, and see the Statue of Libery. N16E wanted to know if they could fit in Ellis Island and a visit to Grand Central Station, and still get a cheesecake for Carnegie Deli. Summershowers responded saying that was a lot to do in two days, and recommended that they drop Macy's and Ellis Island. Summershowers said that John's had the best pizza and recommended that N16E get reservations for Carmine's immediately. Summershowers said that Carmine's was family style and recommended the Flintstone-sized Porterhouse steak. Summershowers said that if N16E goes to the World Trade Center, to stop by St. Paul's Chapel, and to get pastrami a Carnegie Deli as well as a cheesecake. N16E thanked Bettine, and asked for a recommended order in visiting places, and said the Carmine's might be a little too formal. \\
            \\
            \hline
        \end{tabular}
        }
    \caption{A case from FORUM. The conversation's domain is trip. We underline some vital facts in the conversation. \textcolor{red}{Red} denotes incorrect content in the generated summaries. \textcolor{blue}{Blue} indicates what appears in \textsc{TANet}'s summary but is not covered by the ground truth.}
    \label{tab:forum-c1}
\end{table*}

\end{document}